\documentclass{article}

\usepackage[preprint]{neurips_2026}

\usepackage[utf8]{inputenc} % allow utf-8 input
\usepackage[T1]{fontenc}    % use 8-bit T1 fonts
\usepackage{hyperref}       % hyperlinks
\usepackage{url}            % simple URL typesetting
\usepackage{booktabs}       % professional-quality tables
\usepackage{amsfonts}       % blackboard math symbols
\usepackage{nicefrac}       % compact symbols for 1/2, etc.
\usepackage{microtype}      % microtypography
\usepackage{xcolor}         % colors
\usepackage{amsmath}        % math
\usepackage{graphicx}       % figure
\usepackage{tabularx}       % table
\usepackage{float}          % float
\usepackage{placeins}  

\title{Distilling Bayesian Belief States into Language Models for Auditable Negotiation}

\author{%
  Zongqi~Cui \\
  Emory University\\
  \texttt{zongqi.cui@emory.edu} \\
  \And
  Baihan~Lin \\
  Icahn School of Medicine at Mount Sinai\\
  \texttt{baihan.lin@mssm.edu} \\
}

\begin{document}

\maketitle
% Add this after \maketitle:
\begingroup
\renewcommand\thefootnote{}\footnote{Code available at \url{https://github.com/kaneis1/CaSiNo_negotiation-agent}}
\addtocounter{footnote}{-1}
\endgroup

\begin{abstract}
Negotiation agents must infer what their counterpart values, update those beliefs over dialogue turns, and choose actions under uncertainty. End-to-end large language models (LLMs) can imitate negotiation dialogue, but their opponent beliefs are usually implicit and difficult to inspect. We propose BOND (Bayesian Opponent-belief Negotiation Distillation), a framework for auditable negotiation. BOND consists of an LLM-based Bayesian teacher that scores dialogue contexts against the six possible opponent priority orderings, updates a posterior over those orderings, and uses the posterior for menu-based decision making, as well as a smaller 8B student language model that emits both negotiation actions and normalized posterior beliefs as tagged text. In the CaSiNo negotiation dataset, BOND outperforms the state-of-the-art and achieves mean Brier score 0.085 over opponent-priority posteriors. The distilled student preserves much of this belief signal, achieving Brier 0.114, below the uniform six-ordering reference of $5/36 \approx 0.139$. Compared with a 70B structured-CoT baseline, the significantly smaller 8B student model yields substantially better elicited posterior calibration. We further showcase auditability through posterior trajectories, belief-versus-policy error decomposition, and posterior-prefix interventions. These diagnostics reveal that distillation preserves a scoreable belief report more strongly than causal belief-conditioned control, making weak belief-action coupling visible, not hidden.
\end{abstract}

\section{Introduction}
\label{sec:intro}

Negotiation agents must infer what their counterpart values, update that belief over dialogue turns, and act under uncertainty. End-to-end language models can generate fluent negotiation dialogue, but their beliefs about the opponent are usually implicit: when a model accepts, rejects, or counteroffers, it is difficult to tell whether the action follows from a stable opponent model or from shallow dialogue imitation. Bayesian cognitive models provide the complementary strength: they expose priors, likelihoods, and posteriors, but usually require hand-designed likelihood functions that do not scale naturally to free-form negotiation language.

We combine these two views. Our central commitment is that \emph{language supplies the likelihood, and structured inference performs the update}. An LLM scores how compatible a dialogue context is with each candidate opponent priority ordering; a Bayesian module converts these scores into a posterior over hidden opponent preferences; and a menu planner uses the posterior for accept/counteroffer decisions. We then distill this Bayesian teacher into an 8B student that emits the posterior as tagged text alongside its selected intent, content, and utterance. The resulting model remains a generative negotiation agent, but its opponent belief state is externalized and can be scored at every turn. We call this framework \textbf{BOND} (\textbf{B}ayesian \textbf{O}pponent-belief \textbf{N}egotiation \textbf{D}istillation).

We study this framework in the CaSiNo dataset \citep{chawla2021casino}, where two participants negotiate over Food, Water, and Firewood and each privately holds one of six priority orderings. Prior CaSiNo opponent modeling predicts the opponent's final priority order from dialogue prefixes; we instead study the full posterior trajectory over all six orderings. This makes the model auditable in a stronger sense than aggregate prediction accuracy: the exposed posterior lets us distinguish wrong-belief failures from right-belief/wrong-action failures, inspect individual belief trajectories, and test whether posterior correction changes the student's action. 

On 150 held-out CaSiNo dialogues, totaling 1054 turn-level predictions from the \texttt{mturk\_agent\_1} perspective, the Bayesian teacher achieves mean Brier score 0.085 over opponent-priority posteriors. The distilled 8B student preserves much of this belief signal, achieving Brier 0.114, below the uniform six-ordering reference of $5/36 \approx 0.139$. A 70B structured-CoT baseline has stronger accept/reject F1, but its elicited posterior is poorly calibrated, with Brier 0.194. Thus, the main claim is not that the student is the best negotiation policy; it is that a smaller model can expose a valid, scoreable belief interface while retaining usable decision behavior. Contributions and presentations are as following:

% \paragraph{Contributions.}
\begin{itemize}
    \item We formulate negotiation opponent modeling as Bayesian belief-state distillation, using an LM as a likelihood estimator over six opponent priority orderings.
    \item We distill the Bayesian teacher into an 8B student that emits normalized posterior beliefs as tagged text, achieving Brier 0.114 versus a 0.139 uniform reference.
    \item We demonstrate auditability through belief-policy error decomposition, posterior trajectories, and posterior-prefix intervention.
    \item We compare against a 70B structured-CoT baseline and show that elicited posterior calibration remains worse than the student's native posterior.
    \item We provide ablations showing that posterior tags primarily improve inspectability, while causal belief-action coupling remains limited after supervised fine-tuning (SFT) distillation.
\end{itemize}
\section{Related Work}
\label{sec:related}

Automated negotiation is commonly decomposed into bidding, opponent modeling, and acceptance strategy \citep{baarslag2016exploring,chang2020multi}. Neural dialogue work such as Deal-or-No-Deal studies negotiation as end-to-end language generation \citep{lewis2017deal}, while CaSiNo adds a richer campsite scenario with natural-language strategies, private priorities, SVO, Big Five traits, and outcome annotations \citep{chawla2021casino}. Prior CaSiNo opponent modeling predicts the opponent's priority ordering from dialogue prefixes \citep{chawla2022opponent}. We build on this task but replace final-order prediction with a turn-level posterior over all six orderings.

Bayesian cognitive models explain behavior as inference under uncertainty, making priors, likelihoods, and posteriors explicit \citep{griffiths2008bayesian,marr1982vision}. This is attractive for negotiation because the hidden state is concrete: the opponent's private priority ordering. The challenge is that realistic dialogue evidence is free-form and strategic. We therefore use the LM as a likelihood-like language interpreter while keeping the posterior update explicit.

Recent LLM negotiation systems use structured prompting for observation, inference, planning, and response generation \citep{abdelnabi2024cooperation}, and negotiation coaching systems show that LLM feedback can improve human bargaining \citep{shea2024acellmbasednegotiationcoaching}. BEDA similarly treats belief estimates as mediators of strategic dialogue acts \citep{li2025bedabeliefestimationprobabilistic}. Our contribution is to distill an explicit Bayesian posterior into a smaller LM and evaluate that posterior directly with turn-level Brier score and auditability diagnostics.
\section{Methods}
\label{sec:methods}

We design a modular negotiation agent that separates language interpretation, belief revision, and action selection. The central design choice is to treat the language model as a likelihood estimator rather than as a complete black-box policy. At each turn, the agent observes a partial CaSiNo dialogue, scores how compatible the dialogue is with each possible opponent priority ordering, updates a Bayesian posterior over these orderings, and uses the posterior to choose whether to accept, reject, or propose an allocation. Under BOND framework, we then distill this Bayesian teacher into an 8B student that emits the posterior directly as tagged text, making belief auditable at every turn.

\subsection{Task and Hypothesis Space}
\label{sec:methods:task}

We evaluate in the CaSiNo campsite-negotiation domain \citep{chawla2021casino}. Each dialogue involves two participants negotiating over three item types:
\[
\mathcal{I}=\{\textsc{Food}, \textsc{Water}, \textsc{Firewood}\}.
\]
For each participant, the three items are assigned private priorities High, Medium, and Low, worth 5, 4, and 3 points per package, respectively. Since there are three items, the opponent's hidden priority ordering belongs to a six-element hypothesis space:
\[
\Theta = \mathrm{Perm}(\mathcal{I}), \qquad |\Theta| = 3! = 6.
\]
This six-state hypothesis space is deliberately small. It matches CaSiNo's private-priority design and makes turn-level posterior scoring unambiguous, which is useful for a first auditable-belief benchmark. We do not claim that human negotiation preferences are generally discrete or purely ordinal. Larger domains could replace $\Theta$ with continuous priority weights over issues, structured combinatorial preferences, or particle-based posterior approximations.
A belief state is therefore a categorical posterior distribution
\[
p_t(\theta) = P(\theta \mid h_t),
\]
where $h_t$ is the dialogue history available at turn $t$ and $\theta \in \Theta$ is a candidate opponent priority ordering.

\subsection{Architecture}
\label{sec:methods:architecture}

\begin{figure}[t]
    \centering
    % Replace the path below with your generated architecture figure.
    \includegraphics[width=0.95\linewidth]{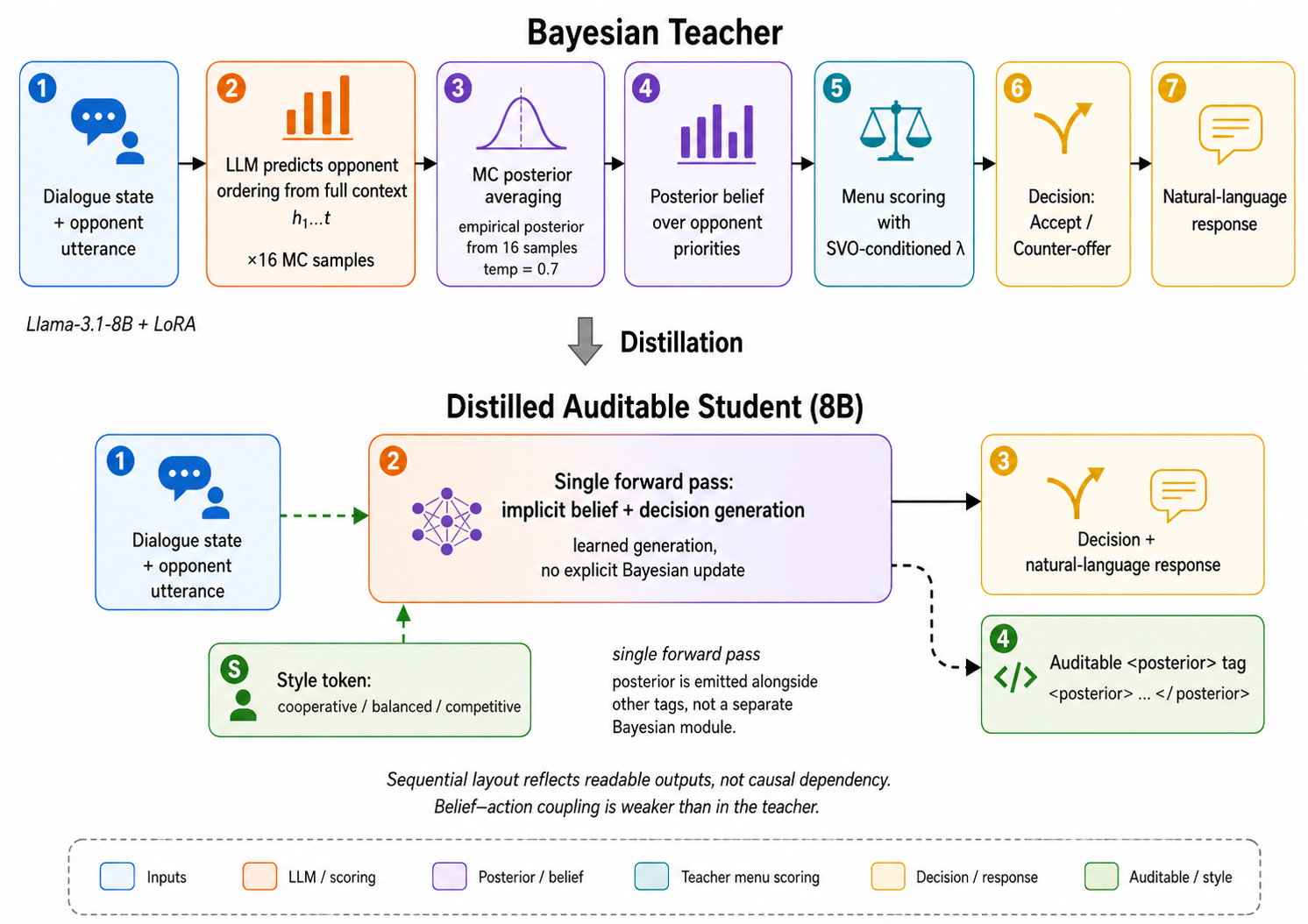}
    \caption{
    \textbf{Modular architecture of the BOND framework}. The LLM supplies likelihood scores over six opponent priority orderings; a Bayesian module updates the posterior; the menu module scores candidate allocations using self-utility and posterior expected opponent utility; the student model is distilled to emit posterior, intent, content, and utterance tags.
    }
    \label{fig:architecture}
\end{figure}

Figure~\ref{fig:architecture} summarizes the architecture. Given a dialogue history $h_t$, the agent proceeds through four stages.

First, the language module scores the compatibility between the observed dialogue and each candidate opponent ordering $\theta \in \Theta$. This produces a likelihood-like signal for each hypothesis. Second, the Bayesian module updates the posterior over the six orderings. Third, the menu module enumerates feasible CaSiNo allocations and scores them using the agent's own utility and the posterior expected opponent utility. Fourth, the decision module selects an action: accept an existing offer, reject and counter-offer, submit a deal, walk away, or produce a non-deal utterance.

This decomposition differs from a standard prompted negotiation agent. A prompted structured-CoT baseline may internally reason about the opponent, but it does not expose a calibrated posterior distribution that can be scored. In contrast, both our teacher and student output explicit posterior beliefs, allowing turn-level evaluation of belief quality.

\subsection{Bayesian Teacher}
\label{sec:methods:teacher}

The Bayesian teacher uses a LoRA-adapted Llama-3.1-8B-Instruct model ~\citep{meta2024llama31modelcard,dubey2024llama} as the likelihood scorer. In the main evaluation, the teacher is evaluated on 150 held-out dialogues from the \texttt{mturk\_agent\_1} perspective, producing 1054 turn-level records. Exact model and adapter details are reported in the reproducibility appendix.

For each turn $t$ and each candidate ordering $\theta \in \Theta$, the teacher obtains a language-based compatibility score. We transform these scores using the fixed evaluation configuration: likelihood temperature 25.0 and clipping to $[-3,3]$. The posterior is then updated as
\begin{equation}
    p_t(\theta)
    =
    \frac{
        \tilde{L}_t(\theta) p_{t-1}(\theta)
    }{
        \sum_{\theta' \in \Theta}
        \tilde{L}_t(\theta') p_{t-1}(\theta')
    },
    \label{eq:bayes-update}
\end{equation}
where $\tilde{L}_t(\theta)$ is the clipped and temperature-transformed likelihood score for ordering $\theta$. Unless otherwise specified, the prior $p_0$ is uniform over the six orderings. We use ``Bayesian'' in the Marr computational-level sense: the model specifies a hypothesis space, a prior, evidence-dependent likelihood-like weights, and a posterior update rule. The transformed score $\tilde{L}_t(\theta)$ is not claimed to be a calibrated generative likelihood of the utterance under $\theta$; it is an LM-derived compatibility function used as the evidential term in a discrete Bayesian update.

In practice, the teacher scores each ordering against the full dialogue context $h_t$ rather than against the incremental utterance $u_t$ alone. An ablation testing per-utterance likelihood decomposition finds that single-utterance likelihoods are poorly calibrated, producing posteriors worse than uniform at every tested temperature (Section~\ref{sec:results:ablation}). The full-context scoring delegates temporal evidence integration to the LLM's attention mechanism, while the Bayesian structure provides the hypothesis space, posterior format, and belief--action interface. 

The teacher uses posterior sampling with $k=16$ samples and posterior temperature 0.7. For action selection, it uses $\lambda=1.0$, accept margin 5, and accept floor 0.5. These values are held fixed across the main teacher evaluation. In the main teacher run, the Bayesian teacher achieves mean Brier score 0.085 over 1054 posterior predictions, with Accept-F1 0.897 and bid cosine 0.744.

\subsection{Distillation to Student}
\label{sec:methods:student}

We distill the Bayesian teacher into a student with the same 8B base model family. The student uses the same 8B base model family with a LoRA adapter trained on teacher-generated trajectories. Exact adapter details are reported in the reproducibility appendix. Distillation data is constructed from teacher trajectories: for each dialogue context, the target output includes the teacher posterior, a selected intent, selected content when applicable, and the natural-language utterance.

The student is trained to emit a tagged intermediate representation:
\begin{verbatim}
<posterior>...</posterior>
<selected_intent>...</selected_intent>
<selected_content>...</selected_content>
<utterance>...</utterance>
\end{verbatim}
The valid intents are \texttt{submit}, \texttt{accept}, \texttt{reject}, \texttt{walkaway}, and \texttt{utter}. The posterior tag contains one normalized probability for each of the six priority orderings. The selected-content tag is either null or a canonical CaSiNo allocation satisfying the item-count constraint that each item type has three total packages.
The exact prompt and output schema is illustrated in Appendix~\ref{app:prompts}, Figure~\ref{fig:student_schema}.
This output format is central to the paper's auditability claim. The model does not merely generate an utterance; it also externalizes the belief distribution that supports its action. During evaluation, malformed outputs are parsed defensively and logged, but the main balanced-student run produced no parse errors: all 1054 calls had valid posterior and intent fields.

We evaluate the balanced style token in the main paper. Earlier style-token and Big Five analyses are treated as null or limitations results rather than main claims, because fixed-style strategy distributions were indistinguishable and human Big Five--strategy correlations were near zero.

\subsection{Menu Scoring}
\label{sec:methods:menu}

Given posterior $p_t(\theta)$, the menu module enumerates all feasible CaSiNo allocations and scores each allocation by
\[
    \mathrm{score}(\pi)
    =
    U_{\mathrm{self}}(\pi)
    +
    \lambda
    \mathbb{E}_{\theta \sim p_t}
    [U_{\mathrm{opp}}(\pi \mid \theta)] .
\]
Here $U_{\mathrm{self}}$ is the agent's utility for its allocated items, and the expectation marginalizes opponent utility over the six posterior orderings. We use $\lambda=1.0$ in the main teacher and audit experiments. SVO-sensitive variants change $\lambda$ only as an appendix diagnostic; because this additive form makes $\lambda$ an opponent-utility exchange rate rather than a bounded self-other interpolation, we do not treat SVO conditioning as a main performance claim.

\subsection{Evaluation Protocol}
\label{sec:methods:eval}

We use Protocol 3, a turn-level replay evaluation over held-out CaSiNo dialogues. Each model observes the dialogue history up to the current turn and produces a structured response. We evaluate on a 150-dialogue held-out subset of CaSiNo, corresponding to approximately 15\% of the full corpus, and score 1054 turn-level predictions from the \texttt{mturk\_agent\_1} perspective. The held-out split is fixed before evaluation, stratified by integrative potential, and split by participant rather than by dialogue.

For each turn, we record four metrics:

\paragraph{Posterior Brier score.}
For models that expose a posterior, we compute Brier score against the true opponent priority ordering:
\begin{equation}
    \operatorname{Brier}(p_t,y)
    =
    \frac{1}{|\Theta|}
    \sum_{\theta \in \Theta}
    \left(p_t(\theta)-\mathbf{1}[\theta=y]\right)^2
    \tag{8}
    \label{eq:brier}
\end{equation}
For $|\Theta|=6$, a uniform posterior has
\[
\frac{1}{6}\left[\left(1-\frac{1}{6}\right)^2 + 5\left(\frac{1}{6}\right)^2\right] = \frac{5}{36} \approx 0.139.
\]
Prompted structured-CoT baselines do not expose a posterior, so Brier score is undefined for them.

\paragraph{Accept prediction.}
When a human turn corresponds to an accept-eligible decision point, we compare the model's accept/reject decision against the human decision and report F1, precision, recall, and accuracy. The evaluation uses accept margin 5 and accept floor 0.5.

\paragraph{Bid similarity.}
When the model emits a concrete offer, we compare the predicted allocation to the human allocation using bid cosine similarity over canonical item-count vectors. Free-text offers are parsed into canonical CaSiNo allocations when possible.

\paragraph{Strategy macro-F1.}
We compare generated utterance strategies against CaSiNo's turn-level strategy annotations. The label set includes small talk, self-need, other-need, no-need, elicit-preference, coordination, fairness appeals, empathy, undervaluing partner, and non-strategic utterances.

The main comparison includes three systems: a 70B structured-CoT live baseline, the Bayesian teacher, and the distilled 8B student. The structured-CoT baseline is evaluated with max-new-tokens 800, temperature 0.3, and seed 2024. The Bayesian teacher is evaluated with max-new-tokens 96 and temperature 0.7. The distilled student is evaluated with temperature 0.0, student max-new-tokens 256, and the balanced style token.

\section{Results}
\label{sec:results}
We organize the results around the central claim: explicit posterior belief states can be transferred through distillation and evaluated turn by turn. We first test posterior calibration, then show how exposed beliefs diagnose failures, and finally use ablations to identify which parts of the architecture carry the belief-tracking signal. Decision metrics are treated as behavioral sanity checks rather than the sole measure of success.

\subsection{Auditable Posterior Quality Transfers Through Distillation}
\label{sec:results:calibration}

We report two opponent-modeling results: the distilled 8B student preserves much of the Bayesian teacher's posterior quality, and Table~\ref{tab:external_main} anchors the 8B model class against prior CaSiNo opponent-modeling baselines.
On the 150-dialogue held-out CaSiNo subset, corresponding to 1054 turn-level predictions from the \texttt{mturk\_agent\_1} perspective, the Bayesian teacher achieves mean Brier score 0.085 over the six opponent-priority hypotheses. 
The distilled student achieves mean Brier score 0.114. 
Both are below the uniform six-way posterior reference of $5/36 \approx 0.139$.

\begin{figure}[t]
    \centering
    \includegraphics[width=0.85\linewidth]{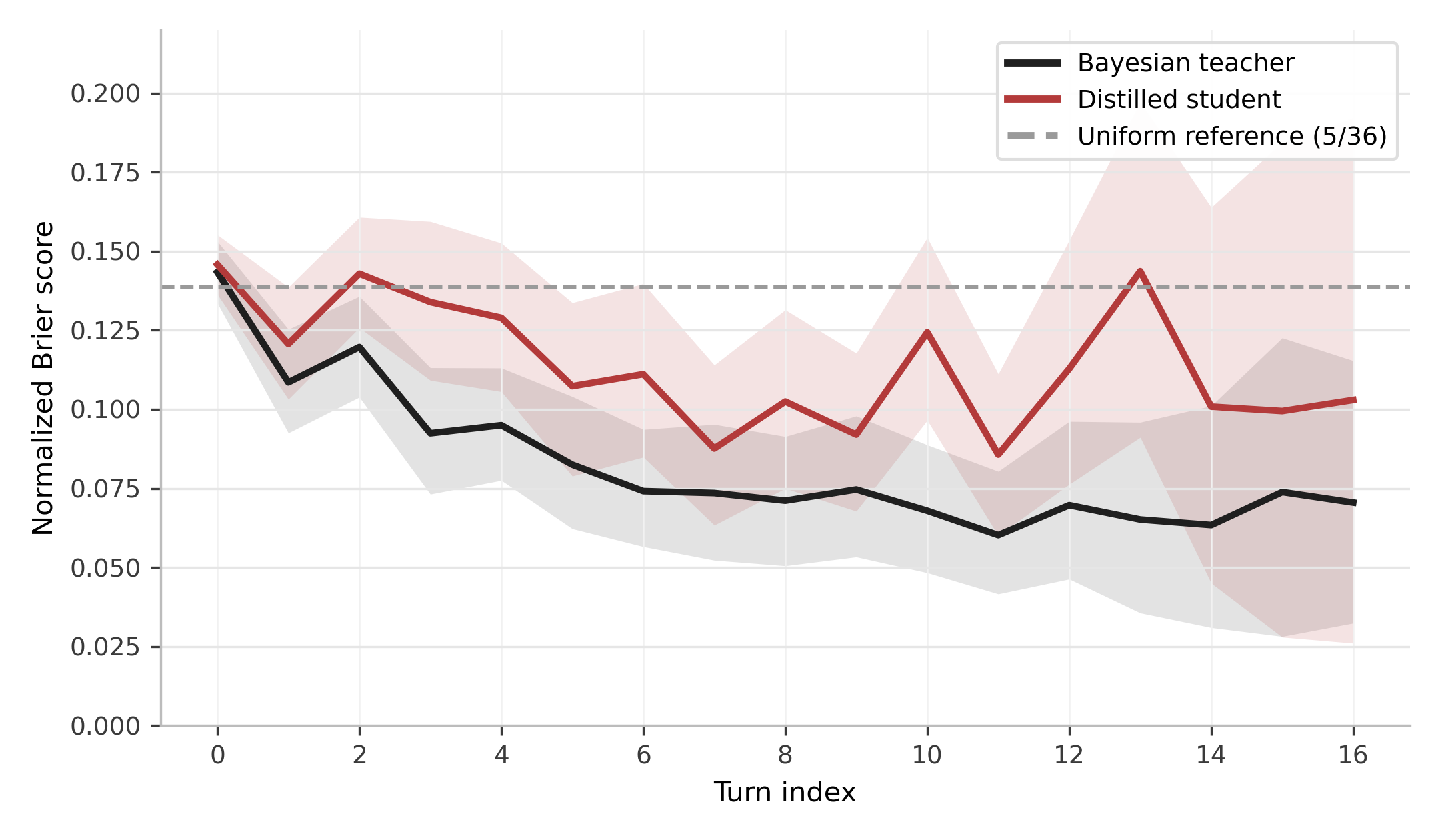}
    \caption{
\textbf{Turn-level posterior Brier score.} On the 150-dialogue held-out CaSiNo subset, shaded bands show 95\% dialogue-level bootstrap CIs from 2000 resamples of held-out dialogues. The dashed line is the uniform six-ordering reference, $5/36 \approx 0.139$. Lower is better. Bands widen at later turns because fewer dialogues contain those turn indices.
}
    \label{fig:brier}
\end{figure}

Figure~\ref{fig:brier} shows the Brier trajectory by turn index. 
For the main plot, we include only turn indices with support $n \geq 10$ for both teacher and student. 
The student's overall mean Brier is 0.114, below the uniform reference of 0.139. At three supported turn indices (0, 2, and 13), the per-turn Brier slightly exceeds the reference, consistent with weak evidence at the beginning of dialogue and lower support at late turns. On the remaining 14 of 17 supported turns, the student remains below the uniform reference.

The structured-CoT baseline does not natively expose a posterior. To enable a direct comparison, we elicit one via self-consistency over the same six orderings, with $n=5$ samples at temperature 0.7 and majority-vote shares as the elicited posterior. The resulting mean Brier score is 0.194 on $n=1054$ matched turn-level records, \emph{above} the uniform reference of 0.139 and substantially worse than both the student's 0.114 and the teacher's 0.085. Even at $5\times$ the inference cost of a single forward pass, the prompted baseline does not match the student's posterior calibration.

A second diagnostic concerns belief--action coupling. On the 679 turns where the baseline emitted a parseable structured offer, the offer matched the utility-maximizing allocation under the elicited MAP ranking on only 31.1\% of turns under a strict criterion (offer equals the top split under $U_{\mathrm{self}} + 1.0 \cdot U_{\mathrm{opp}}$) and 26.2\% under a loose criterion (offer gives the opponent more units of higher-ranked than lower-ranked issues). The remaining 375 of 1054 turns produced unstructured natural-language utterances and are excluded from this rate. The Bayesian teacher does not have this gap by construction: the posterior is the direct input to menu scoring. The distilled student preserves this output format, though Section~\ref{sec:results:ablation} reveals that belief--action coupling is weaker after distillation. This dissociation between elicited belief and action in the prompted baseline motivates the hybrid architecture.

\begin{table}[t]
\centering
\small
\begin{tabular}{lccc}
\toprule
Model & EMA $\uparrow$ & Top-1 $\uparrow$ & NDCG@3 $\uparrow$ \\
\midrule
BoW-Ranker~\citep{chawla2022opponent} & 27.71 & 52.98 & 64.31 \\
BERT CD+CA+DND~\citep{chawla2022opponent} & 44.22 & 69.21 & 76.03 \\
RoBERTa CD+CA+DND~\citep{chawla2022opponent} & 48.72 & 70.03 & 77.14 \\
70B structured-CoT prompted & 37.30 & 62.99 & 70.94 \\
BOND - Supervised SFT 8B & \textbf{53.84} & \textbf{76.21} & \textbf{80.88} \\
\bottomrule
\end{tabular}
\caption{
\textbf{External CaSiNo opponent-modeling comparison.} Under the k-penalty protocol of \citet{chawla2022opponent}, prior rankers predict opponent orderings but do not emit actions or calibrated turn-level belief states; full protocol details are in Appendix~\ref{app:external}.
}
\label{tab:external_main}
\end{table}

\subsection{Auditability in Use: Separating Belief Errors from Policy Errors}
\label{sec:results:auditability}

Aggregate posterior metrics show whether the student is calibrated, but auditability requires a stronger test: can the exposed belief state explain why a negotiation decision succeeds or fails? We therefore inspect the distilled student's held-out turns by comparing three objects: the emitted posterior, the emitted action, and the deterministic Bayesian menu recommendation induced by that posterior.

Across 1054 held-out student turns, the student has MAP accuracy 0.518, mean confidence 0.725, mean entropy 0.783 bits, and mean class-normalized Brier 0.114. We identify 181 audit-supported decision turns, where the turn contains a formal decision, a structured bid, or an accept-eligible pending offer. For each such turn, we compute whether the posterior MAP matches the true opponent ordering, whether the student's action agrees with the human behavioral reference, and whether the action aligns with the deterministic menu recommendation under the student's own posterior.

Table~\ref{tab:audit_decomp} decomposes the resulting errors. The main pattern is that belief quality and action quality can fail independently: 67 turns have the correct MAP belief but are inconsistent with the menu planner, while 27 turns have a wrong MAP belief but are internally consistent with the wrong posterior. This separation is the core auditability benefit: an observed failure can be assigned to belief tracking, downstream policy selection, or weak coupling between the two.

Appendix~\ref{app:auditability_cases} visualizes posterior trajectories and lists selected audit cases. In dialogue 33, the student's observed action matches the human accept decision, but its MAP belief swaps Water and Firewood; under this wrong posterior, the deterministic menu would reject and counteroffer, whereas under the correct posterior it would accept. In dialogue 533, the student has the correct MAP belief, Water $>$ Firewood $>$ Food, but rejects a deal that both the human and the menu planner accept. In dialogue 627, the wrong MAP belief changes the menu recommendation itself, producing a bid aligned with the wrong-posterior menu but not with the correct-posterior menu.
\begin{table}[t]
\centering
\small
\begin{tabular}{lcc}
\toprule
Belief state & Menu-aligned action & Menu-inconsistent action \\
\midrule
MAP correct & 49 & 67 \\
MAP wrong   & 27 & 38 \\
\bottomrule
\end{tabular}
\caption{
\textbf{Belief-policy decomposition on 181 audit-supported decision turns from the distilled student}. MAP correctness measures whether the exposed posterior's highest-probability ordering matches the true opponent priority ordering. Menu alignment measures whether the emitted action matches the deterministic Bayesian menu recommendation induced by the student's own posterior. The table shows that belief errors and action-selection errors are separable.
}
\label{tab:audit_decomp}
\end{table}

Finally, we test whether intervening on the emitted posterior changes the student's action. Correct-prefix intervention replaces the posterior tag with the ground-truth one-hot posterior before the student generates the remaining fields; adversarial-prefix intervention replaces it with an incorrect one-hot posterior. Correct-prefix intervention changes the student's action or bid on only 1.5\% of turns, and adversarial-prefix intervention changes it on only 1.8\%. In the joined changed cases, correct-prefix agreement with the human improves in 5 cases but worsens in 11. We therefore do not interpret posterior correction as a reliable control mechanism. Instead, it exposes a limitation: SFT distillation transfers the posterior report and the action behavior more strongly than it transfers the causal posterior$\rightarrow$planner dependency of the Bayesian teacher.

\subsection{Ablation Analysis}
\label{sec:results:ablation}

We evaluate nine ablation conditions to isolate the 
contributions of individual architectural components. 
Table~\ref{tab:ablation} reports the results; all 
conditions use the same 150-dialogue held-out split 
and Protocol~3 evaluation.

\begin{table}[t]
\centering
\small
\begin{tabular}{lcccc}
\toprule
Variant & Brier $\downarrow$ & MAP $\uparrow$ & Accept-F1 $\uparrow$ & Bid cos $\uparrow$ \\
\midrule
\multicolumn{5}{l}{\textit{Reference systems}} \\
MC teacher (main) & 0.085 & --- & 0.897 & 0.744 \\
Full student (ablation rerun) & 0.112 & 0.522 & 0.908 & 0.909 \\
Uniform reference & 0.139 & 0.167 & --- & --- \\
\midrule
\multicolumn{5}{l}{\textit{A1: Is MC sampling necessary?}} \\
Direct zero-shot & 0.163 & 0.126 & 0.926 & 0.717 \\
Direct SFT (ground-truth) & 0.163 & 0.453 & 0.890 & 0.743 \\
\midrule
\multicolumn{5}{l}{\textit{A3: Does the LLM beat rules?}} \\
Rule-based likelihood & 0.134 & 0.191 & 0.886 & 0.723 \\
\midrule
\multicolumn{5}{l}{\textit{A4: Does the posterior tag help decisions?}} \\
Action-only student & --- & --- & 0.921 & 0.926 \\
\midrule
\multicolumn{5}{l}{\textit{A2d: Is the posterior causally used?}} \\
Adversarial prefix & 0.333 & 0.000 & 0.886 & 0.942 \\
Correct prefix (sanity) & 0.000 & 1.000 & 0.894 & 0.923 \\
\midrule
\multicolumn{5}{l}{\textit{A5: Does the posterior help planning?}} \\
Uniform posterior planner & 0.139 & 0.124 & 0.926 & 0.733 \\
\midrule
\multicolumn{5}{l}{\textit{A8: Incremental Bayes diagnostic}} \\
Incremental, $T{=}100$ & 0.176 & 0.128 & 0.815 & 0.727 \\
Incremental, $T{=}25$ & 0.223 & 0.128 & 0.815 & 0.727 \\
\bottomrule
\end{tabular}
\caption{
\textbf{Ablation results on 150 held-out CaSiNo dialogues ($n{=}1054$ turns)}. Bid cosine supports vary by condition. The adversarial and correct prefix rows inject a forced posterior into the student's generation; action change rate is 1.8\% (see Appendix~\ref{app:coupling} for the full diagnostic protocol). The ablation rerun uses the same checkpoint evaluated with a different random seed; minor differences from Table~\ref{tab:decision_quality} reflect sampling variance.
}
\label{tab:ablation}
\end{table}
\paragraph{Full-context MC posterior estimation is better calibrated than direct posterior baselines.}
The direct zero-shot posterior (A1a) achieves Brier 0.163, well above the uniform reference of 0.139, confirming that the base LoRA model cannot produce calibrated posteriors without MC sampling. More strikingly, a model trained directly on ground-truth smoothed labels (A1b) achieves high MAP accuracy (45.3\%) but identical Brier (0.163). This model learns to identify the correct ordering but does not produce calibrated uncertainty — it is overconfident on wrong predictions and underconfident on correct ones. The MC teacher achieves better calibration (Brier 0.085) without any ground-truth labels, indicating that the sampling-based posterior provides a calibration advantage over supervised prediction. These results suggest that the sampled full-context posterior is better calibrated than direct posterior prediction under our tested baselines.
\paragraph{The LLM contributes language understanding beyond keyword matching.}
The rule-based likelihood (A3) achieves Brier 0.134, below the uniform reference, confirming that simple pattern matching on CaSiNo's structured preference discourse provides genuine signal. However, it remains substantially worse than the MC teacher (0.085), indicating that the LLM extracts evidence from indirect and ambiguous utterances that keyword rules cannot capture.
\paragraph{The posterior tag aids auditability, not decision accuracy.}
The action-only student (A4), trained without the posterior tag, achieves Accept-F1 0.921 and bid cosine 0.926 — slightly higher than the full student on both metrics. This confirms that the posterior tag's contribution is inspectability rather than decision quality. The adversarial prefix test (A2d) reveals a related finding: injecting an adversarial posterior at generation time changes the student's action on only 1.8\% of turns. The student has learned to emit accurate posteriors and accurate actions through parallel pathways rather than through the sequential posterior→planner chain present in the teacher. Enforcing tighter belief-action coupling through auxiliary losses or constrained decoding is an important target for future distillation.
\paragraph{The incremental Bayesian decomposition does not work.}
To test whether per-utterance likelihood decomposition contributes to posterior quality, we implement an incremental Bayes provider that scores only the newest opponent utterance and applies Equation~\ref{eq:bayes-update} with a temporal prior. Across all tested likelihood temperatures (1–100) and clipping ranges, the incremental posterior is worse than uniform (best Brier 0.176 at T=100), with MAP accuracy below chance (12.8\%). The posterior degrades as evidence accumulates, indicating that single-utterance likelihoods from a dialogue-trained model are not calibrated as conditional probabilities. The working MC teacher succeeds because the LLM integrates evidence across the full dialogue context in each forward pass — temporal integration is performed by the model's attention mechanism, not by Bayesian multiplication. Full temperature and clipping sweep results are reported in Appendix~\ref{app:sensitivity}.

\subsection{Decision and Diagnostic Checks}
\label{sec:results:decision}

Table~\ref{tab:decision_quality} shows that auditability does not make the student the strongest observable policy: the 70B structured-CoT baseline has higher Accept-F1 (0.947 vs. 0.908). The student's advantage is different: it exposes a valid posterior on every turn, with Brier 0.114 versus 0.194 for the elicited 70B posterior. Bid cosine is conditional on sparse structured-bid emission ($n=14$), so we treat it as descriptive rather than as evidence of robust bid coverage.

\begin{table}[H]
\centering
\small
\begin{tabular}{lcccc}
\toprule
Model & Accept-F1 $\uparrow$ & Bid cosine $\uparrow$ & Strategy macro-F1 $\uparrow$ & Brier $\downarrow$ \\
\midrule
70B structured-CoT & \textbf{0.947} & 0.815 {\scriptsize ($n=29$)} & 0.160 & 0.194$^{\ddagger}$ \\
Bayesian teacher & 0.897 & 0.744 {\scriptsize ($n=86$)} & 0.048 & \textbf{0.085} \\
Distilled 8B student & 0.908 & \textbf{0.915} {\scriptsize ($n=14$)} & \textbf{0.190} & 0.114 \\
\bottomrule
\end{tabular}
\caption{
\textbf{Protocol 3 turn-level evaluation on 150 held-out CaSiNo dialogues}. Bid cosine is computed only on turns with parseable native structured bids; supports are shown in parentheses. $^{\ddagger}$The 70B baseline does not natively expose a posterior; its Brier is computed from majority-vote shares over $n=5$ self-consistency samples at temperature 0.7. Teacher and student Brier are native single-pass posteriors.
}
\label{tab:decision_quality}
\end{table}

As a limited cross-domain sanity check, zero-shot renamed transfer to Deal-or-No-Deal obtains DND-normalized Brier 0.1335 versus 0.3190 for direct posterior prompting, and few-shot adaptation improves to 0.1166. DND Brier uses a sum-normalized multiclass definition with uniform reference $1/6$, whereas the main CaSiNo Brier uses class-mean normalization with uniform reference $5/36$; the values are therefore comparable only within their respective sections. As an SVO diagnostic, binary accept decisions are insensitive to matched versus mismatched SVO labels, while allocation behavior is highly sensitive to the scale of $\lambda$. Thus, SVO-as-$\lambda$ is a scale-sensitive diagnostic rather than evidence that personality conditioning improves negotiation decisions. Full SVO and DND details are in Appendices~\ref{app:svo_expanded} and~\ref{app:dnd_transfer}.

\section{Conclusion.}
We presented BOND, a Bayesian belief-state distillation framework for auditable negotiation. An LLM-based teacher produces posterior beliefs over opponent priority orderings, and an 8B student learns to emit these beliefs as tagged text alongside negotiation actions. The student preserves much of the teacher's posterior calibration and exposes a scoreable belief interface, but our audit diagnostics show that standard SFT preserves belief reports more strongly than causal posterior-to-action coupling. Future auditable agents should therefore move beyond belief reporting toward constrained decoding, auxiliary planning losses, or explicit posterior-conditioned policies.

\bibliographystyle{plainnat}
\bibliography{references}
\appendix

\FloatBarrier
\section{Limitations and Broader Impact}
\label{app:limitations}

\paragraph{Parameter scale of the supervised baseline.}
Our supervised SFT baseline uses an 8B foundation model (Llama-3.1-8B)  ~\citep{meta2024llama31modelcard,dubey2024llama} 
with LoRA, while \citet{chawla2022opponent} use BERT-base and 
RoBERTa-base ($\sim$125M parameters). The comparison reflects current-generation supervised 
performance on opponent-priority prediction at 8B scale rather than a 
parameter-matched evaluation; we report it to establish that 8B models 
reach competitive performance on this task before evaluating the 
additional contribution of our Bayesian framework. The 70B prompted 
baseline, evaluated under the matched k-penalty protocol, demonstrates that 
scale alone without supervised structure does not match the supervised 8B 
or the smaller RoBERTa ranker on this task, despite outperforming the 
BoW-Ranker baseline.
\paragraph{Scope of evaluation.}
Our experiments are limited to the CaSiNo campsite-negotiation domain and to a 150-dialogue held-out split evaluated primarily from the \texttt{mturk\_agent\_1} perspective. CaSiNo provides a useful controlled setting because the opponent's hidden priority ordering belongs to a six-element hypothesis space, but this structure is much simpler than open-ended real-world negotiation. The results may not transfer directly to multi-party, multi-round, or high-stakes negotiations with larger action spaces, richer preference structures, or institutional constraints.

\paragraph{Teacher and posterior quality.}
The distilled student inherits both the strengths and errors of the Bayesian teacher. The teacher's likelihood model is based on LLM-scored compatibility between dialogue contexts and six candidate priority orderings, rather than on ground-truth cognitive likelihoods. The posterior should therefore be interpreted as an auditable computational belief state, not as a direct measurement of a human negotiator's mental state.

\paragraph{Decision-policy limitations.}
The distilled student preserves posterior quality but does not dominate the 70B structured-CoT baseline on all decision metrics. In particular, the structured-CoT baseline has stronger accept/reject F1. Therefore, our claim is not that the distilled model is the best negotiation policy. Our goal is not to win every negotiation metric; it is to make the hidden opponent model of a negotiation agent explicit, distillable, and scoreable at every turn. This limitation is also the intended use case for auditability: because the belief state is exposed, weak belief--action coupling can be detected rather than hidden inside an end-to-end policy.

\paragraph{Core modeling limitations.}
The current system has three important limitations. First, the opponent hypothesis space contains only the six ordinal priority orderings in CaSiNo. This makes posterior scoring clean, but does not cover continuous issue weights or larger combinatorial preference spaces. Second, the student preserves posterior quality more reliably than structured bid coverage: only 14 of 1054 turns contain native parseable structured bids in the main run. Improving bid emission is therefore a clear target for future distillation, for example by upweighting bid-emitting turns or adding an auxiliary bid-format loss. Third, the Bayesian teacher's strategy macro-F1 is near floor because the teacher is not trained as a surface strategy imitator; it is a belief-update and menu-scoring model. These limitations narrow the behavioral claims but do not undermine the central contribution: a distilled negotiation agent can externalize a turn-level posterior that is validly parsed and directly scoreable.

\paragraph{SVO operationalization.}
Our SVO analysis uses an additive utility function in which $\lambda$ acts as an opponent-utility exchange rate. This differs from a convex self-other interpolation and makes the numerical scale of $\lambda$ important. As a result, our SVO findings should be read as a methodological sensitivity analysis rather than as evidence that SVO matching improves negotiation decisions.

\paragraph{Potential societal impact.}
Negotiation agents could support education, training, and decision assistance by making belief states and tradeoffs explicit. However, the same methods could also be misused to build more persuasive or manipulative bargaining systems. We therefore emphasize posterior auditability, explicit uncertainty reporting, and careful evaluation of social-orientation parameters as safeguards against opaque or overconfident deployment.

\paragraph{Perspective Replication}
A further limitation is that the main Protocol 3 results are reported from the \texttt{mturk\_agent\_1} perspective; full \texttt{mturk\_agent\_2} replication remains future work.
\FloatBarrier
\section{Dataset Split and Quality Filtering}
\label{app:dataset_quality}

We begin from the full CaSiNo corpus of 1030 dialogues. The final split contains 880 training dialogues and 150 held-out test dialogues. The held-out split is fixed before evaluation, stratified by integrative potential, and split by participant rather than by dialogue to reduce participant leakage.

For quality-filtered distillation, we compute participant-level quality-audit rows and style-specific subsets. The quality audit contains 1722 participant-level rows. For each style weight $w \in \{0.2,0.5,0.8\}$, the style-filtered subset contains 516 selected examples. The three style weights correspond to cooperative, balanced, and competitive selection regimes. The audit file \texttt{quality\_audit.csv} records the underlying quality scores and filtering decisions; qualitative checks are summarized in \texttt{validation\_notes.md}.
\FloatBarrier
\section{Distillation Data Audit}
\label{app:distill_audit}

The final teacher-distillation corpus contains 10293 rows generated from Bayesian teacher trajectories, with a generation failure rate of 0.0. After intent-aware oversampling, the student training split contains 10713 rows, and the held-out student evaluation split contains 991 rows.

We audit the distillation data by style, intent, and perspective. The style-conditioned rows are derived from the three quality-filtered subsets described in Appendix~\ref{app:dataset_quality}. The valid intent labels are \texttt{submit}, \texttt{accept}, \texttt{reject}, \texttt{walkaway}, and \texttt{utter}. Intent repeat factors are used during oversampling to reduce collapse toward the majority \texttt{utter} class and to preserve rare structured decision outputs.
\begin{table}[H]
\centering
\small
\begin{tabular}{lr}
\toprule
Quantity & Count \\
\midrule
Teacher-distillation rows & 10293 \\
Teacher generation failure rate & 0.0 \\
Post-oversampling train rows & 10713 \\
Evaluation rows & 991 \\
\bottomrule
\end{tabular}
\caption{Distillation data audit summary from \texttt{day7\_summary.json} and \texttt{day8\_data\_summary.json}.}
\label{tab:distill_audit}
\end{table}
\FloatBarrier
\section{Structured Output Validity}
\label{app:output_validity}

The auditability claim depends on the student reliably emitting parseable belief states. In the full 150-dialogue balanced-student evaluation, the student produced valid posteriors for 1054/1054 turn-level calls and valid intents for 1054/1054 calls, with 0 parse errors. Across student variants, parse-failure logs are empty. We also ran the repository unit tests in the project environment; all 47 tests passed.

\begin{table}[H]
\centering
\small
\begin{tabular}{lcc}
\toprule
Field & Valid / Total & Failure rate \\
\midrule
Posterior & 1054 / 1054 & 0.0 \\
Intent & 1054 / 1054 & 0.0 \\
Parse errors & 0 / 1054 & 0.0 \\
Unit tests & 47 / 47 passed & -- \\
\bottomrule
\end{tabular}
\caption{Structured output validity audit from \texttt{turn\_eval\_student\_balanced\_full150/turn\_summary.json}.}
\label{tab:structured_validity}
\end{table}

\FloatBarrier
\section{Full Metric Breakdown}
\label{app:full_metrics}

Table~\ref{tab:decision_quality} in the main paper reports compressed headline metrics. Here we provide the supporting metric breakdown, including accept confusion matrices, bid supports, per-label strategy F1, and Brier-by-turn support.

The Brier trajectory uses supported turn indices 0--16. Across these supported turns, the student's maximum normalized Brier score is 0.1458, slightly above the uniform six-ordering reference of $5/36 \approx 0.139$, occurring at turn index 0 where little dialogue evidence is yet available.
\begin{table}[H]
\centering
\small
\begin{tabular}{lcc}
\toprule
Statistic & Teacher & Student \\
\midrule
Supported turn indices & 0--16 & 0--16 \\
Minimum support threshold & $n \geq 10$ & $n \geq 10$ \\
Mean normalized Brier & 0.085 & 0.114 \\
Maximum supported-turn Brier & -- & 0.1458 \\
Uniform reference & 0.139 & 0.139 \\
\bottomrule
\end{tabular}
\caption{Brier-by-turn support summary from \texttt{headline\_numbers.csv} and \texttt{headline\_checks.json}.}
\label{tab:brier_support}
\end{table}
\FloatBarrier
\section{Bid Coverage Diagnostic}
\label{app:bid_coverage}

The distilled student has high native bid cosine when it emits a parseable structured bid, but bid emission is sparse. In the balanced-student run, native bid cosine is 0.915 on $n=14$ native structured bids. The student produced 22 predicted bid turns, and coverage relative to gold submit turns is 0.163. A manual spotcheck found that 4/5 cases in which the student stayed in utterance mode were conservative-correct: the student avoided forcing a structured bid when the dialogue context did not clearly support one.

This diagnostic explains why bid cosine is treated as conditional bid quality rather than as a robust bid-coverage metric in the main paper.
\FloatBarrier
\section{Protocol Alignment: Replay vs. Turn-Level Evaluation}
\label{app:protocol_alignment}

We distinguish earlier Protocol-1 replay numbers from the final Protocol-3 turn-level evaluation. Protocol 1 replays a baseline trajectory and can lose support when the baseline terminates early. In the alignment audit, Protocol-1 replay covered only 28/87 accept-decision turns because the baseline often terminated before later human decision points. Protocol 3 instead evaluates models at matched turn-level contexts, giving matched support across systems. Therefore, the final paper uses Protocol 3 for the headline baseline comparison.
\FloatBarrier
\section{Expanded SVO Diagnostics}
\label{app:svo_expanded}

The main paper reports the compact SVO match-vs-mismatch diagnostic and summarizes the scale confound. Here we provide expanded subgroup analyses, including first-offer integrativeness, accept-F1 by SVO group, and $\lambda$-change diagnostics. These tables support the main claim that SVO-as-$\lambda$ under additive scoring decomposes social-label effects from numerical scale effects.
\FloatBarrier
\section{Big Five and Style-Token Null Results}
\label{app:bigfive}

We tested whether Big Five-conditioned style tokens produced distinguishable strategy distributions across cooperative, balanced, and competitive variants. They did not: the fixed-style strategy distributions were statistically indistinguishable, $\chi^2 p=0.998$. A separate human-data analysis found Big Five $\times$ strategy correlations to be near zero, with maximum $|r|=0.069$.

\begin{figure}[h]
    \centering
    \includegraphics[width=\linewidth]{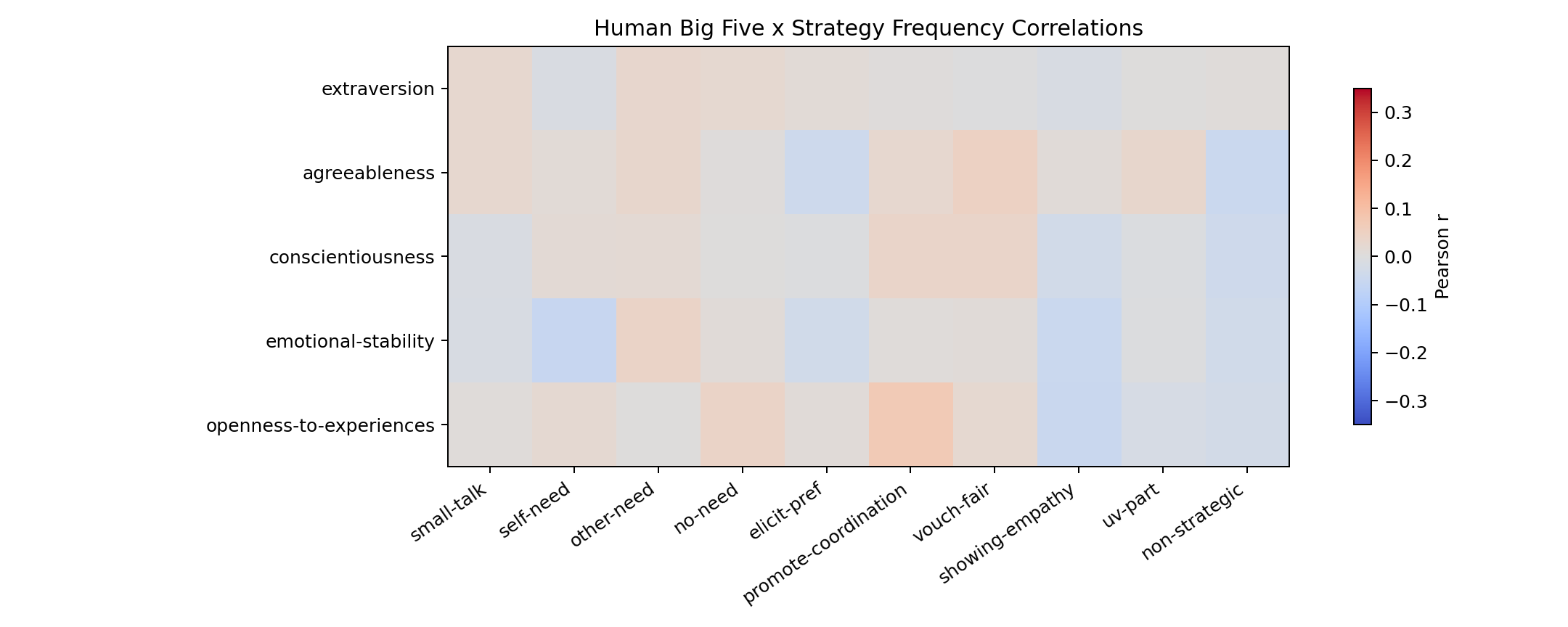}
    \caption{Human Big Five $\times$ strategy Pearson correlation heatmap. Correlations are uniformly small, with maximum absolute correlation $|r|=0.069$.}
    \label{fig:bigfive_heatmap}
\end{figure}
\FloatBarrier
\section{Prompt and Output Schema}
\label{app:prompts}

\begin{figure}[h]
    \centering
    \includegraphics[width=\linewidth]{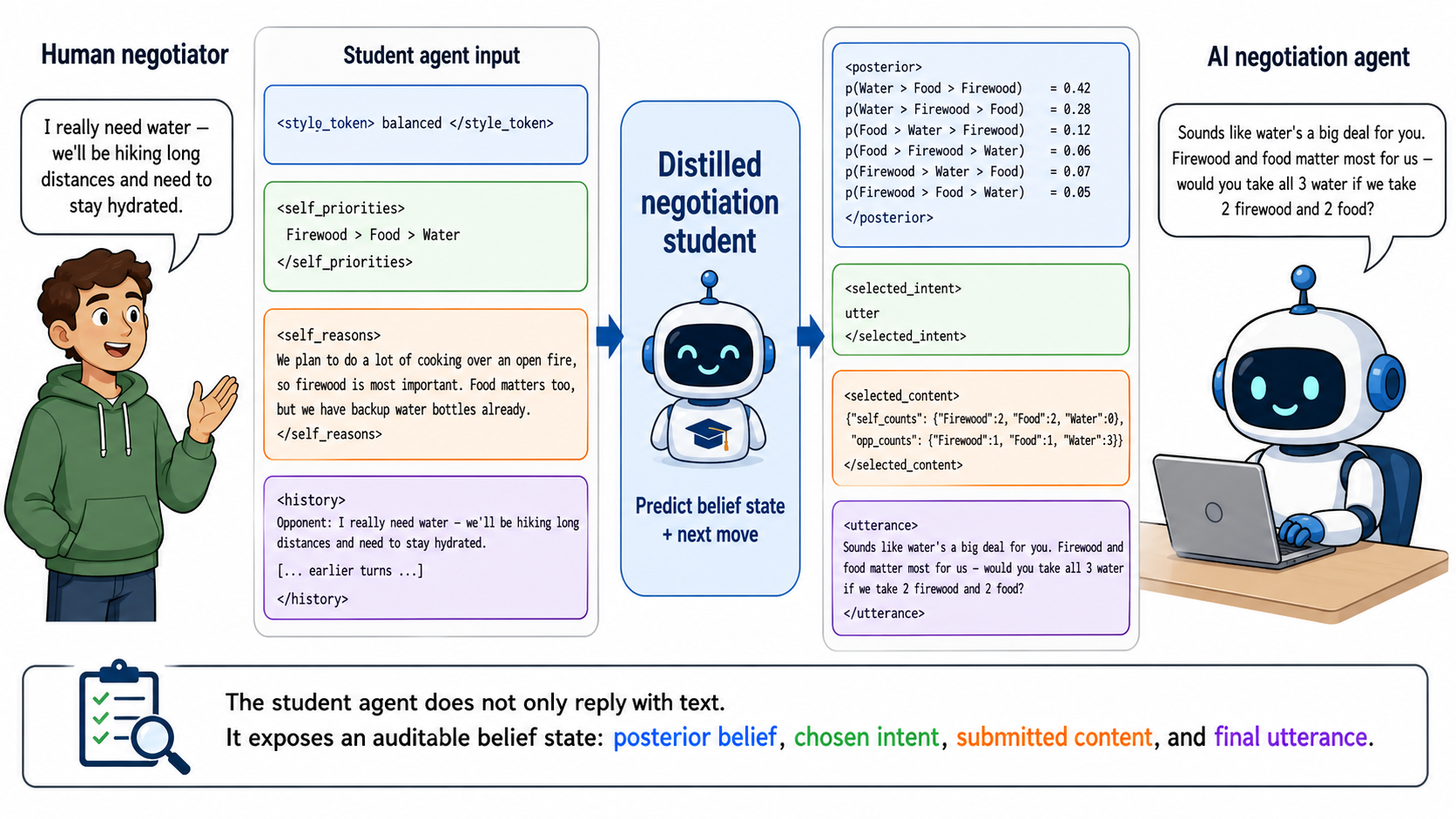}
    \caption{Student prompt and output schema for one turn. The student does not only generate a response; it externalizes an auditable belief state through tagged posterior, intent, content, and utterance fields.}
    \label{fig:student_schema}
\end{figure}

The student is required to emit four tagged fields:
\begin{verbatim}
<posterior>...</posterior>
<selected_intent>...</selected_intent>
<selected_content>...</selected_content>
<utterance>...</utterance>
\end{verbatim}
Figure~\ref{fig:student_schema} illustrates the student input-output format for one CaSiNo turn. The student receives a style token, the speaker's private priorities and reasons, and the dialogue history; it emits a posterior over six opponent priority orderings, a selected intent, selected content, and a final utterance.

\FloatBarrier
\section{LoRA Training Details}
\label{app:lora_training}

The distilled student is trained with LoRA rank $r=16$, $\alpha=32$, dropout 0.05, learning rate $10^{-4}$, batch size 4, gradient accumulation 4, maximum sequence length 1536, and seed 42. The best evaluation loss is 1.018. LoRA target modules and the full training arguments are reported in the released \texttt{sft\_run\_meta.json} and \texttt{training\_args.json} files.

\begin{table}[H]
\centering
\small
\begin{tabular}{lr}
\toprule
Hyperparameter & Value \\
\midrule
LoRA rank & 16 \\
LoRA alpha & 32 \\
LoRA dropout & 0.05 \\
Learning rate & $1\times 10^{-4}$ \\
Batch size & 4 \\
Gradient accumulation & 4 \\
Max sequence length & 1536 \\
Seed & 42 \\
Best eval loss & 1.018 \\
\bottomrule
\end{tabular}
\caption{Student LoRA training configuration from \texttt{sft\_run\_meta.json} and \texttt{training\_args.json}.}
\label{tab:lora_training}
\end{table}

The posterior contains one normalized probability for each of the six candidate opponent priority orderings. The likelihood prompt asks the teacher model to score the compatibility between the dialogue context and each candidate ordering. The structured-CoT baseline prompt decomposes negotiation into observation, opponent inference, planning, and response generation stages. Complete prompt files are included in the supplementary material.
\FloatBarrier
\section{Reproducibility Statement}
\label{app:reproducibility}

All main experiments use the same 150-dialogue held-out CaSiNo split and Protocol 3 turn-level evaluation. Each model observes the dialogue history up to a target turn and produces a structured response. We report the number of evaluated turn records and metric supports for accept decisions, bid cosine, strategy macro-F1, and Brier score.

\paragraph{Systems.}
The main systems are:
\begin{itemize}
    \item \textbf{Structured-CoT baseline:} a 70B prompted structured-CoT negotiation baseline evaluated under Protocol 3.
    \item \textbf{Bayesian teacher:} a Llama-3.1-8B-Instruct model  ~\citep{meta2024llama31modelcard,dubey2024llama} with LoRA used as a likelihood scorer, followed by an explicit Bayesian posterior update and menu-scoring policy.
    \item \textbf{Distilled student:} a Llama-3.1-8B-Instruct student~\citep{meta2024llama31modelcard,dubey2024llama} with LoRA trained to emit posterior, intent, selected content, and utterance tags.
\end{itemize}

\paragraph{Main hyperparameters.}
For the Bayesian teacher, we use posterior sampling with $k=16$, posterior temperature 0.7, likelihood temperature 25.0, likelihood clipping to $[-3,3]$, accept margin 5, and accept floor 0.5. The main teacher uses $\lambda=1.0$ unless otherwise specified. The distilled student is evaluated with temperature 0.0, maximum 256 new tokens, and the balanced style token. The structured-CoT baseline is evaluated with maximum 800 new tokens, temperature 0.3, and seed 2024.

\paragraph{Result artifacts.}
The main results are produced from final evaluation artifacts: the 70B structured-CoT baseline, the Bayesian teacher, and the balanced distilled student. The Brier trajectory, main metrics table, SVO subgroup summaries, and SVO accept diagnostics are generated from the same held-out split. Code, run scripts, split identifiers, and final result artifacts are available at \url{https://github.com/kaneis1/CaSiNo_negotiation-agent}.
\FloatBarrier
\section{Compute Resources}
\label{app:compute}

Training and evaluation were run on an institutional HPC cluster. The main model sizes were 70B for the structured-CoT baseline and 8B for the Bayesian teacher and distilled student.

\begin{table}[H]
\centering
\small
\scriptsize
\setlength{\tabcolsep}{3pt}
\begin{tabularx}{\linewidth}{l l X X}
\toprule
Run & Hardware & Approx. time & Output \\
\midrule
70B structured-CoT baseline & 2$\times$ NVIDIA H100 80GB HBM3  & 6h 26m 57s & Protocol 3 metrics \\
Bayesian teacher evaluation & 1$\times$ NVIDIA H100 PCIe 80GB  & 30m 35s & posterior + menu records \\
8B LoRA distillation & 1$\times$ NVIDIA H100 PCIe 80GB  & 1h 54m 39s & student checkpoint \\
8B student evaluation & 1$\times$ NVIDIA H100 PCIe 80GB & 1h 39m 14s & Protocol 3 metrics \\
\bottomrule
\end{tabularx}
\caption{Compute resources used for the main experiments. Times are LSF wall-clock run times for the final successful jobs. The 8B student evaluation row reports the uncached generation run; a subsequent cache-only rerun took 1m 37s on 1$\times$ NVIDIA H100 80GB HBM3.}
\label{tab:compute}
\end{table}

\FloatBarrier
\section{Assets, Licenses, and Data Use}
\label{app:assets}

We use the CaSiNo dataset and annotations from \citet{chawla2021casino}. CaSiNo is a previously published negotiation-dialogue dataset; we do not collect new human-subject data. We use the dataset for research evaluation and derived train/test splits, and any released split files or processed artifacts will preserve the original dataset attribution and access conditions.

For cross-domain transfer, we use the Deal-or-No-Deal negotiation dataset introduced by \citet{lewis2017deal}. We use it only for the strict-ordering transfer diagnostic reported in Appendix~\ref{app:dnd_transfer}. The original dataset source is credited in the references, and any derived transfer artifacts will document the preprocessing used to map item values to strict six-way orderings.

The Bayesian teacher and distilled student use Llama-3.1-8B-Instruct~\citep{meta2024llama31modelcard,dubey2024llama} as the base model, and the structured-CoT baseline uses Llama-3.3-70B-Instruct~\citep{meta2024llama33modelcard,dubey2024llama}. We use these models under the applicable Meta Llama Community License and acceptable-use policy. For Llama-3.1, the applicable license is the Llama 3.1 Community License Agreement. If we release LoRA adapter checkpoints or other derivative model artifacts, the release will include the required Llama license copy, NOTICE attribution, and documentation stating that the artifact is built with Llama. We will not redistribute full base-model weights.

Our derived assets include train/test split identifiers, quality-filtered training subsets, evaluation logs, posterior trajectories, result tables, Brier trajectory plots, auditability case studies, SVO diagnostic summaries, and LoRA adapter checkpoints. Any released code or artifacts will include a README documenting data provenance, preprocessing, evaluation commands, known limitations, and the licenses or access terms governing the original datasets and model dependencies. The released repository or supplementary package will also include a license file specifying terms for code and derived artifacts.

\FloatBarrier
\section{LLM Usage}
\label{app:llm_usage}

LLMs are central to the method. The Bayesian teacher uses an LLM as a likelihood scorer over six candidate opponent priority orderings. The distilled student is itself an LLM trained to emit posterior beliefs, selected intent, selected content, and utterance text. We also compare against a 70B structured-CoT LLM baseline.

LLMs were not used as a substitute for human-subject data collection in this paper. The human dialogue data and annotations come from the previously published CaSiNo dataset. The LLM components are used for modeling, inference, distillation, and evaluation within the computational pipeline described in the Methods section.
\FloatBarrier
\section{Cross-Domain Transfer to Deal-or-No-Deal}
\label{sec:results:dnd}

To test whether the belief-updating interface is specific to CaSiNo, we evaluate it on Deal-or-No-Deal (DND), a separate multi-issue negotiation dataset in which agents bargain over books, hats, and balls. Unlike CaSiNo, DND item values are not fixed by a 5/4/3 priority scale and may contain ties. We therefore evaluate belief transfer only on strict partner orderings, where the opponent's values map cleanly to one of six item-rank permutations. Because DND dialogues are short, we emphasize EMA@2 and k-penalized metrics over $k=1$--3.

\begin{table}[t]
\centering
\small
\setlength{\tabcolsep}{4pt}
\begin{tabular}{lcccc}
\toprule
Model & Brier $\downarrow$ & EMA@2 $\uparrow$ & EMA$_{1:3}$ $\uparrow$ & NDCG$_{1:3}$ $\uparrow$ \\
\midrule
Direct posterior, renamed & 0.3190 & 0.1806 & 0.1743 & 0.5170 \\
Rule baseline & 0.1605 & 0.2369 & 0.2259 & 0.5675 \\
Bayesian transfer, native & 0.1438 & 0.4796 & 0.4827 & 0.7852 \\
Bayesian transfer, renamed & 0.1335 & 0.5126 & 0.5029 & 0.7969 \\
Few-shot, native empirical & \textbf{0.1166} & \textbf{0.5981} & \textbf{0.5879} & \textbf{0.8541} \\
Few-shot, renamed empirical & 0.1207 & 0.5845 & 0.5694 & 0.8459 \\
\bottomrule
\end{tabular}
\caption{
Cross-domain transfer to Deal-or-No-Deal on strict partner orderings ($n=1695$ turn snapshots).
DND Brier uses the sum-normalized multiclass definition, with six-way uniform reference $1/6 \approx 0.167$.
EMA@2 is emphasized because DND dialogues are short; EMA$_{1:3}$ and NDCG$_{1:3}$ use the $k=1$--3 penalty window.
}
\label{tab:dnd_transfer}
\end{table}

Table~\ref{tab:dnd_transfer} shows a clear transfer pattern. Direct posterior prediction fails to generalize: the renamed direct posterior baseline obtains Brier 0.3190, substantially worse than the six-way uniform reference of $1/6 \approx 0.167$. A rule baseline is better calibrated than direct prompting, with Brier 0.1605, but remains weak on exact ordering accuracy. In contrast, the Bayesian transfer pipeline already beats the uniform reference in zero-shot transfer, reaching Brier 0.1438 with native DND item names and 0.1335 after lexical renaming. This suggests that the structured posterior interface transfers better than direct posterior prompting.

Few-shot adaptation further improves belief tracking. The native empirical few-shot variant achieves the best posterior quality, with Brier 0.1166, EMA@2 0.5981, and NDCG$_{1:3}$ 0.8541. These results support a limited generalization claim: the method is not merely a CaSiNo-specific text classifier; its six-way hidden-preference inference interface can transfer to another closed-domain bargaining task when the target domain still has a small discrete preference-ordering structure.

\FloatBarrier
\section[SVO-as-lambda Reveals a Scale Confound in Personality-Conditioned Negotiation]{SVO-as-$\lambda$ Reveals a Scale Confound in Personality-Conditioned Negotiation}
\label{sec:results:svo}

SVO-conditioned menu scoring can produce misleading conclusions unless the utility form is specified. In our additive planner, $\lambda$ behaves as an opponent-utility exchange rate rather than as a bounded self-other interpolation parameter, so social-orientation labels and numerical utility-scale effects must be decomposed.

We evaluate whether SVO-conditioned menu scoring changes the Bayesian teacher's behavior on the 150-dialogue held-out set. A key interpretive detail is that our implementation uses the additive menu score from Section~\ref{sec:methods:menu}, rather than a convex self-other interpolation. In a convex formulation, $\lambda$ is bounded and directly describes a tradeoff along the self-other utility segment: moving $\lambda$ changes which side of the segment receives more weight. This is closer to the usual interpretation of SVO angle measures, which summarize a person's relative orientation toward self and other outcomes \citep{murphy2011measuring}. In our additive planner, $\lambda$ instead acts as an exchange rate that converts expected opponent utility into self-utility units. This lets $\lambda > 1$ serve as an altruistic boundary condition, but it also means numerical SVO $\lambda$ values are scale-dependent: small values can be overwhelmed by 3--5 point self-utility increments, while large values can saturate into full concession.

We therefore compare two regimes: a moderate rescaled mapping, with proself $=0.2$ and prosocial $=0.6$, and the original legacy boundary mapping, with proself $=1.0$ and prosocial $=2.0$. 
For each mapping, we also run a swapped-$\lambda$ counterfactual in which proself and prosocial participants receive the opposite $\lambda$. 
This makes the match-vs-mismatch comparison a direct intervention rather than a post-hoc subgroup split.

\paragraph{Accept decisions are insensitive to SVO label assignment within a fixed $\lambda$ regime.}

Table~\ref{tab:svo_accept_null} reports the SVO match-vs-mismatch accept diagnostic. 
Under the rescaled mapping, matched and mismatched SVO assignments produce identical Accept-F1, 0.768 in both conditions. 
Under the legacy boundary mapping, the matched condition is only 0.007 F1 higher than the mismatched condition, 0.911 versus 0.904, and this difference is not significant, Welch $p=0.840$. 
We therefore do not claim that SVO matching improves accept decisions in either regime.

\begin{table}[t]
\centering
\small
\begin{tabular}{lcccc}
\toprule
SVO--$\lambda$ mapping & Matched F1 & Mismatched F1 & $\Delta$ & Welch $p$ \\
\midrule
Rescaled $(0.2, 0.6)$ & 0.768 & 0.768 & 0.000 & 1.000 \\
Legacy $(1.0, 2.0)$ & 0.911 & 0.904 & 0.007 & 0.840 \\
\bottomrule
\end{tabular}
\caption{
SVO match-vs-mismatch accept diagnostic. 
Matching SVO labels to the dialogue does not improve accept prediction under either $\lambda$ mapping. 
We therefore treat SVO-as-$\lambda$ as a sensitivity analysis rather than as a positive conditioning result.
}
\label{tab:svo_accept_null}
\end{table}

\paragraph{Allocation behavior is highly sensitive to $\lambda$ scale.}

The accept-decision null does not mean that $\lambda$ has no behavioral effect. 
Instead, allocation behavior changes strongly with the absolute scale of $\lambda$. 
Under the legacy mapping, proself allocations remain closer to human self-points, but prosocial allocations collapse into excessive concession: prosocial agent self-points are 0.000 in every integrative-potential tercile. 
Under the rescaled mapping, full-concession collapse is avoided and overall joint-points remain close to the human baseline, but the agent over-prioritizes self-points. 
Overall, the rescaled agent achieves 36.505 joint-points compared with 36.493 for humans, but its self-points are 35.202 compared with 18.113 for \texttt{mturk\_agent\_1} humans.

\begin{table}[t]
    \centering
    \small
    \begin{tabular}{llrrrr}
    \toprule
    SVO & IP & Human self & Agent self & Human joint & Agent joint \\
    \midrule
    proself & low  & 17.70 & 13.51 & 34.87 & 36.00 \\
    proself & mid  & 17.65 & 17.42 & 35.96 & 38.36 \\
    proself & high & 20.00 & 21.21 & 39.22 & 41.75 \\
    prosocial & low  & 16.25 & 0.00 & 33.40 & 36.00 \\
    prosocial & mid  & 18.32 & 0.00 & 37.40 & 36.00 \\
    prosocial & high & 18.53 & 0.00 & 37.43 & 36.00 \\
    \bottomrule
    \end{tabular}
    \caption{
    SVO $\times$ integrative-potential breakdown on the 150-dialogue test set under the legacy mapping, $\lambda \in \{1,2\}$. 
    Human values are from the \texttt{mturk\_agent\_1} perspective. 
    The legacy prosocial setting saturates into full concession, while proself allocations remain closer to the human self-point range.
    }
    \label{tab:svo_ip}
\end{table}

\paragraph{Behavioral fidelity to humans on the joint-points null.}

A useful behavioral-fidelity result is that neither humans nor the conditioned agent show a simple prosocial joint-points advantage. 
On the held-out split, human prosocial joint-points are 36.347, while human proself joint-points are 36.653. 
Thus, the human data itself does not support the naive prediction that prosocial orientation should necessarily produce higher joint outcomes in this split. 
The agent's null effect should therefore not be interpreted as a failure to reproduce a robust human prosocial advantage; rather, it is consistent with the absence of that advantage in this evaluation population.

\paragraph{Decomposing the dissociation.}

The strongest supported claim is not that SVO matching improves accept decisions. 
It is that SVO-as-$\lambda$ under additive scoring decomposes two effects: binary accept decisions are insensitive within a fixed $\lambda$ regime, while allocation behavior is sensitive to the absolute scale of $\lambda$. 
This suggests that future SVO-conditioned negotiation agents should distinguish two questions that are often conflated: whether a social-orientation label is matched to a participant, and whether the numerical utility weight induced by that label is calibrated to the scale of the decision problem.
We report Big Five/style-token null checks in Appendix~\ref{app:bigfive}; neither fixed-style strategy distributions nor human Big Five--strategy correlations were load-bearing in the current system.

\FloatBarrier
\section{SVO Subgroup Analysis Under the Rescaled Mapping}
\label{app:rescaled_subgroup}

The main SVO table reports the legacy mapping, $\lambda \in \{1.0,2.0\}$, because it illustrates the saturation failure most clearly. Here we summarize the complementary rescaled mapping, with proself $\lambda=0.2$ and prosocial $\lambda=0.6$.

Under the rescaled mapping, the match-vs-mismatch accept diagnostic remains null: matched and mismatched SVO assignments both achieve Accept-F1 of 0.768. This supports the main-text claim that binary accept decisions are insensitive to SVO label assignment within the tested $\lambda$ regimes. However, allocation behavior still changes with the numerical scale of $\lambda$. Overall, the rescaled agent obtains joint-points close to the human baseline, 36.505 versus 36.493, but over-prioritizes self-points, 35.202 versus 18.113 for \texttt{mturk\_agent\_1} humans.

\begin{table}[H]
\centering
\small
\begin{tabular}{lrrrr}
\toprule
Group & Human self & Agent self & Human joint & Agent joint \\
\midrule
proself & 18.42 & 36.00 & 36.65 & 36.00 \\
prosocial & 17.85 & 34.40 & 36.35 & 37.01 \\
classified all & 18.13 & 35.18 & 36.50 & 36.52 \\
overall & 18.11 & 35.20 & 36.49 & 36.50 \\
\bottomrule
\end{tabular}
\caption{
SVO subgroup results under the rescaled mapping, proself $\lambda=0.2$ and prosocial $\lambda=0.6$. The rescaled mapping avoids the full-concession collapse seen in the legacy prosocial condition, but it over-prioritizes self-points while preserving joint-points near the human baseline.
}
\label{tab:svo_rescaled}
\end{table}

\FloatBarrier
\section{Likelihood Sensitivity Sweeps}
\label{app:sensitivity}

To test whether per-utterance Bayesian updating could work under different hyperparameters, we sweep likelihood temperature $T \in \{1, 5, 10, 25, 50, 100\}$ and clipping range $[-c, c]$ for $c \in \{1, 3, 5, 10, \infty\}$ under the incremental Bayes provider. All runs use the same 150-dialogue held-out split ($n{=}1054$ turns), posterior temperature 
0.7, and $\lambda{=}1.0$.

\begin{table}[H]
\centering
\small
\begin{tabular}{lcccc}
\toprule
Setting & Brier $\downarrow$ & MAP $\uparrow$ & Entropy & Accept-F1 $\uparrow$ \\
\midrule
\multicolumn{5}{l}{\textit{Temperature sweep (clip $[-3,3]$)}} \\
$T{=}1$   & 0.230 & 0.124 & 0.834 & 0.797 \\
$T{=}5$   & 0.230 & 0.124 & 0.833 & 0.797 \\
$T{=}10$  & 0.230 & 0.124 & 0.831 & 0.797 \\
$T{=}25$  & 0.223 & 0.128 & 0.955 & 0.815 \\
$T{=}50$  & 0.202 & 0.128 & 1.323 & 0.815 \\
$T{=}100$ & 0.176 & 0.128 & 1.824 & 0.815 \\
\midrule
\multicolumn{5}{l}{\textit{Clipping sweep ($T{=}25$)}} \\
$[-1,1]$         & 0.201 & 0.124 & 1.329 & 0.815 \\
$[-3,3]$         & 0.223 & 0.128 & 0.955 & 0.815 \\
$[-5,5]$         & 0.223 & 0.128 & 0.955 & 0.815 \\
$[-10,10]$       & 0.223 & 0.128 & 0.955 & 0.815 \\
No clip          & 0.223 & 0.128 & 0.955 & 0.815 \\
\midrule
\textit{References} \\
Uniform          & 0.139 & 0.167 & 2.585 & --- \\
MC teacher       & 0.085 & ---   & ---   & 0.897 \\
\bottomrule
\end{tabular}
\caption{
Incremental Bayes sensitivity sweep. Brier improves monotonically with temperature (approaching uniform), and MAP accuracy remains below chance (16.7\%) at every setting. Clipping ranges $[-3,3]$ through no-clip produce identical results, indicating that single-utterance log-likelihood scores never exceed $\pm 3$. The tighter $[-1,1]$ clip improves Brier only by forcing the posterior closer to uniform. These results confirm that per-utterance likelihood decomposition fails: the incremental provider's posteriors degrade as evidence accumulates rather than improving.
}
\label{tab:sensitivity}
\end{table}

\begin{figure}[h]
    \centering
    \includegraphics[width=0.75\linewidth]{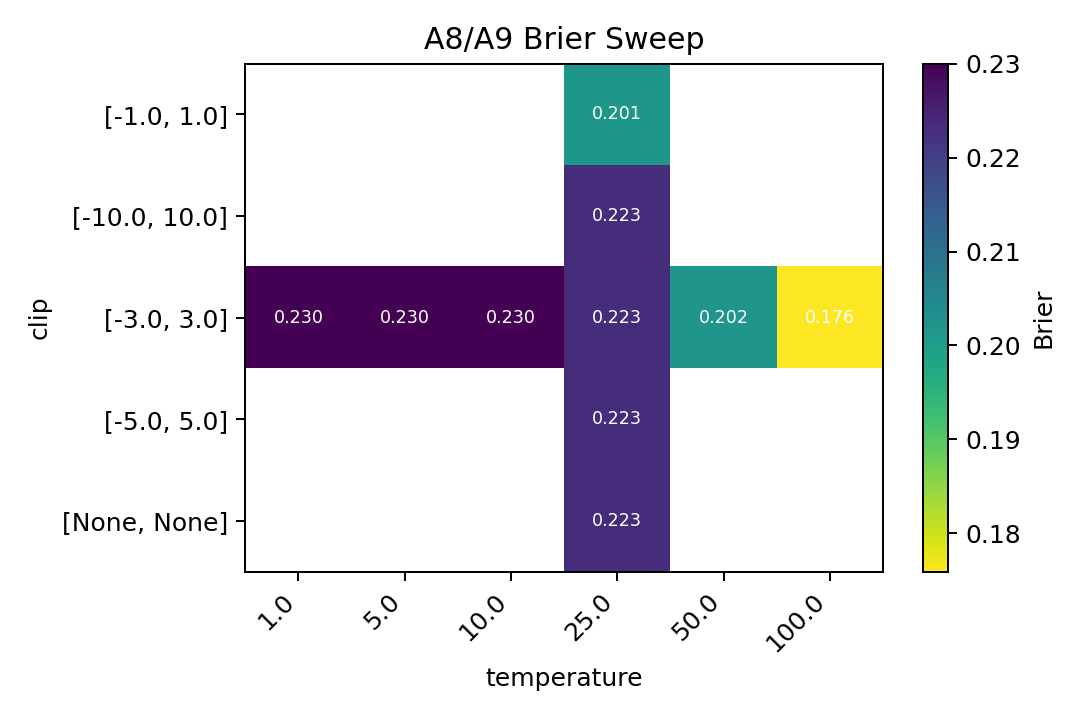}
    \caption{
    Brier score heatmap over likelihood temperature ($T$, rows) and clipping range ($[-c,c]$, columns) for the incremental Bayes provider. The dashed contour marks the uniform reference at 0.139. All cells are above this reference: no combination of temperature and clipping produces posteriors better than chance. Lower temperature and wider clipping produce the worst calibration (dark cells, upper right), while high temperature and tight clipping approach but do not reach the uniform baseline 
    (light cells, lower left).
    }
    \label{fig:sensitivity_heatmap}
\end{figure}

Figure~\ref{fig:sensitivity_heatmap} visualizes the joint effect of temperature and clipping on incremental Bayes posterior quality. The heatmap confirms two patterns visible in Table~\ref{tab:sensitivity}: (1)~Brier improves monotonically with temperature because higher temperature pushes the posterior toward uniform, not because discrimination improves; and (2)~clipping has no effect above $[-3,3]$ because single-utterance log-likelihood scores are already within that range. The only clip setting that changes Brier is $[-1,1]$, which forcibly truncates the scores and produces a more diffuse posterior. No setting in the swept grid produces a posterior better than the uniform reference of $5/36 \approx 0.139$, confirming that per-utterance likelihood decomposition fails on this task regardless of hyperparameter tuning.

\FloatBarrier
\section{Belief--Action Coupling Diagnostic}
\label{app:coupling}

Section~\ref{sec:results:ablation} reports that adversarial posterior injection changes the student's action on only 1.8\% of turns. Here we provide the full diagnostic protocol and breakdown.

\paragraph{Protocol.}
For each of the 1054 evaluation turns, we generate the student's output under three prefix conditions: (1)~\textit{correct prefix}, in which the \texttt{<posterior>} tag is pre-filled with the ground-truth one-hot posterior before the student generates the remaining tags; (2)~\textit{adversarial prefix}, in which the posterior is set to a one-hot distribution concentrated on the \emph{incorrect} ordering farthest from the true one; and (3)~\textit{no prefix} (the baseline full-student run). The student then auto-regressively generates \texttt{<selected\_intent>}, \texttt{<selected\_content>}, and \texttt{<utterance>} conditioned on the injected posterior.

\paragraph{Results.}
Table~\ref{tab:coupling} summarizes the prefix-injection diagnostic.
\begin{table}[H]
\centering
\small
\begin{tabular}{lccccc}
\toprule
Prefix & Brier & MAP & Accept-F1 & Bid cos & Action changed \\
\midrule
No prefix (baseline) & 0.112 & 0.522 & 0.908 & 0.909 & --- \\
Correct prefix       & 0.000 & 1.000 & 0.894 & 0.923 & 1.5\% \\
Adversarial prefix   & 0.333 & 0.000 & 0.886 & 0.942 & 1.8\% \\
\bottomrule
\end{tabular}
\caption{
Belief--action coupling under posterior injection. The Brier and MAP columns reflect the injected posterior, not the student's own beliefs. Action changed reports the fraction of turns where the student's generated intent or selected content differs from the no-prefix baseline. Both correct and adversarial injection change actions on fewer than 2\% of turns, indicating that the student's decisions are driven by learned context--action mappings rather than by the emitted posterior.
}
\label{tab:coupling}
\end{table}

\paragraph{Interpretation.}
The student emits accurate posteriors (Brier 0.112) and accurate decisions (Accept-F1 0.908), but through parallel learned pathways rather than through the sequential posterior$\to$planner chain present in the teacher. The teacher has belief--action coupling by construction: the posterior is the direct input to menu scoring. The student, trained via SFT on the teacher's full tagged output, has learned to reproduce both the posterior and the action but does not enforce consistency between them at generation time.

This dissociation is not a failure of posterior quality --- the student's Brier score confirms that the emitted beliefs are well-calibrated. Rather, it reveals that current SFT distillation transfers each output field independently without preserving the causal dependency structure of the teacher. Future work should explore constrained decoding (e.g., feeding the emitted posterior to an explicit menu scorer before generating the action tags) or auxiliary training losses that penalize posterior-inconsistent actions.

\FloatBarrier
\section{Auditability Case Studies}
\label{app:auditability_cases}

Table~\ref{tab:audit_cases} summarizes the selected audit cases. These cases were generated from the held-out balanced-student records using the audit script. The selector prioritizes turns where the menu recommendation under the student's posterior differs from the menu recommendation under the correct posterior, because these are the turns where inspecting the belief state changes the diagnosis.

\begin{table}[H]
\centering
\scriptsize
\setlength{\tabcolsep}{3pt}
\begin{tabularx}{\linewidth}{l l l l X}
\toprule
Dialogue & Belief status & Student action & Menu contrast & Interpretation \\
\midrule
33, turn 13 & Wrong MAP & Accept & Wrong-posterior menu rejects; correct-posterior menu accepts & Correct observed action but hidden planner risk \\
533, turn 11 & Correct MAP & Reject & Both student-posterior and correct-posterior menus accept & Right belief, wrong action \\
627, turn 12 & Wrong MAP & Bid & Wrong-posterior and correct-posterior menus recommend different bids & Belief error changes menu recommendation \\
\bottomrule
\end{tabularx}
\caption{
Selected auditability cases. The posterior allows us to distinguish harmless observed actions, policy failures under correct beliefs, and belief errors that change the planner recommendation.
}
\label{tab:audit_cases}
\end{table}

\begin{figure}[t]
    \centering
    \includegraphics[width=0.95\linewidth]{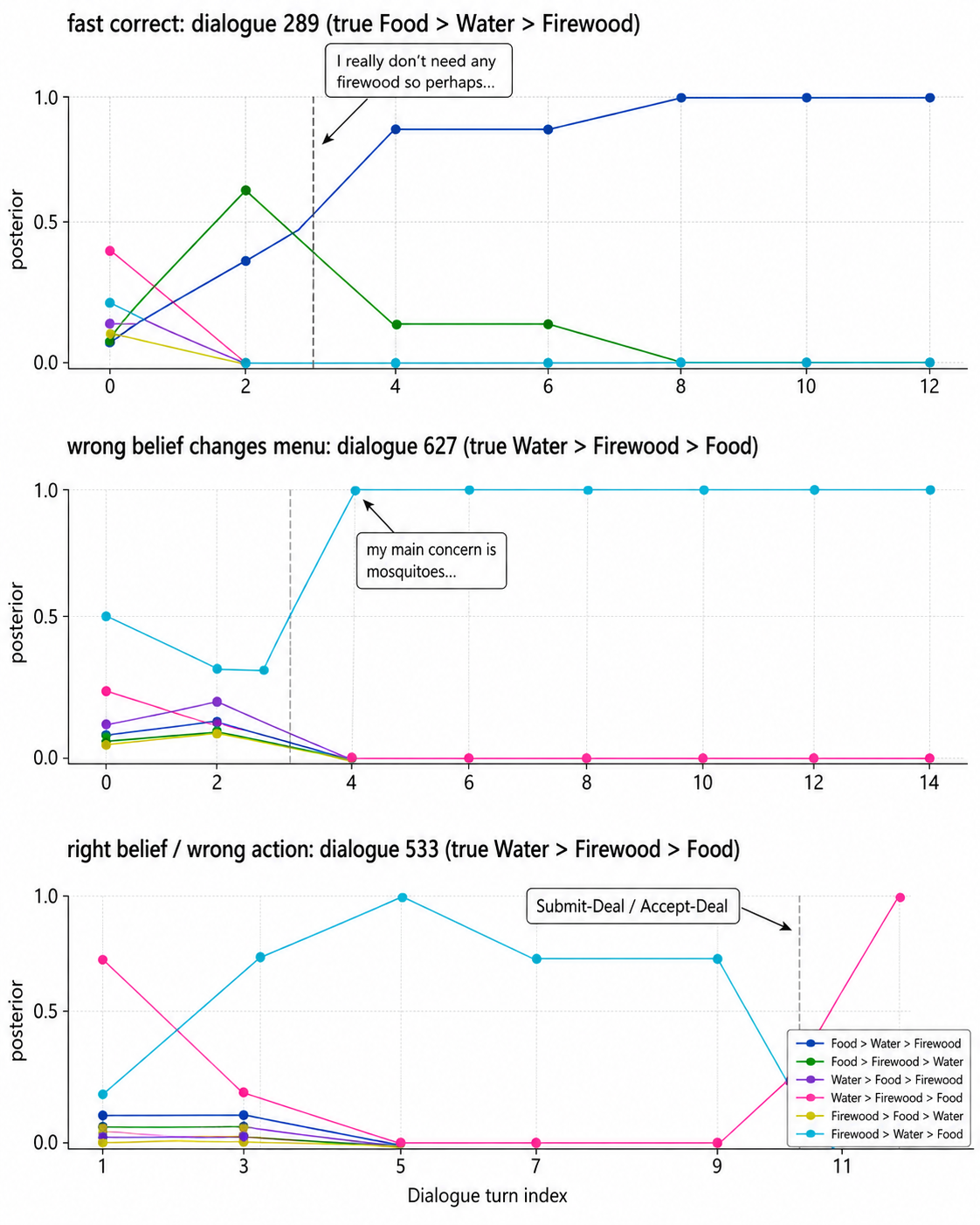}
    \caption{
        Posterior trajectories for three held-out CaSiNo dialogues selected from the distilled student audit. Each line corresponds to one of the six opponent priority orderings. The panels illustrate three auditability patterns: rapid convergence to the correct belief, confident convergence to a wrong belief that changes the menu recommendation, and correct belief at the decision point paired with a wrong action. These trajectories make belief revision inspectable beyond aggregate Brier score.
    }
    \label{fig:audit_trajectories}
\end{figure}

\FloatBarrier
\section{External Comparison to Prior CaSiNo Opponent Modeling}
\label{app:external}

To anchor our results to prior work and to a strong prompted baseline, we 
report opponent priority-ordering prediction across three reference points: 
prior CaSiNo SOTA from \citet{chawla2022opponent}, a 70B prompted 
structured-CoT baseline given the full dialogue, and our supervised SFT 8B 
trained on partial dialogues under 5-fold cross-validation matching the 
protocol of \citet{chawla2022opponent}.\footnote{Following 
\citet{chawla2022opponent}, k-penalty is computed as a linearly weighted 
average over $k=1..5$ with normalized weights $(5,4,3,2,1)/15$, assigning 
higher weight to earlier opponent-observation settings.}
Our protocol matches: 206 unique held-out dialogues per fold and two perspectives 
per dialogue, giving 412 perspective-level evaluation examples per fold, 
identical to the per-fold evaluation count in \citet{chawla2022opponent}.

Table~\ref{tab:opponent_comparison} reports the comparison. The supervised 8B 
model achieves k-penalty Top-1 $76.21 \pm 1.03$ and NDCG@3 $80.88 \pm 0.85$ 
across 5 folds, exceeding the strongest configuration reported by 
\citet{chawla2022opponent} (RoBERTa CD+CA+DND: Top-1 $70.03 \pm 1.63$, NDCG@3 
$77.14 \pm 1.38$) by 6.2 and 3.7 points respectively, with tighter standard 
deviations across folds. The 70B prompted structured-CoT baseline, re-evaluated under the same k-penalty 
protocol over $k=1..5$ on the 150-dialogue held-out split, achieves k-penalty 
EMA 37.30\%, Top-1 62.99\%, and NDCG@3 70.94\%. This places the prompted 70B 
above the BoW-Ranker baseline but well below both BERT/RoBERTa and the supervised 
SFT 8B, indicating that opponent-priority prediction benefits substantially from 
supervised structure, and that scale alone does not close the gap.

\begin{table}[t]
\centering
\small
\begin{tabular}{lccc}
\toprule
Model & EMA $\uparrow$ & Top-1 $\uparrow$ & NDCG@3 $\uparrow$ \\
\midrule
Random {\scriptsize \citep{chawla2022opponent}} & 16.59 (1.22) & 33.99 (1.13) & 49.76 (0.75) \\
BoW-Ranker {\scriptsize \citep{chawla2022opponent}} & 27.71 (1.24) & 52.98 (1.97) & 64.31 (1.67) \\
70B structured-CoT prompted$^{\dagger}$ & 37.30 & 62.99 & 70.94 \\
BERT CD+CA+DND {\scriptsize \citep{chawla2022opponent}} & 44.22 (1.82) & 69.21 (2.05) & 76.03 (1.60) \\
RoBERTa CD+CA+DND {\scriptsize \citep{chawla2022opponent}} & 48.72 (2.03) & 70.03 (1.63) & 77.14 (1.38) \\
\textbf{Supervised SFT 8B (ours)} & \textbf{53.84 (1.58)} & \textbf{76.21 (1.03)} & \textbf{80.88 (0.85)} \\
\bottomrule
\end{tabular}
\caption{
Opponent priority-ordering prediction on CaSiNo (k-penalty metric over 
$k=1..5$). Higher is better. EMA and Top-1 in percentage; NDCG@3 scaled 
to 0--100. Numbers from \citet{chawla2022opponent} are mean (std) over 
5-fold CV as reported in their Table 1, k-penalty column. Our supervised 
SFT 8B numbers are mean (std) over 5-fold CV with matched protocol 
(206 unique held-out dialogues per fold $\times$ 2 perspectives = 412 
perspective-level evaluation examples per fold). $^{\dagger}$The 70B 
prompted baseline (Llama-3.3-70B-Instruct~\citep{meta2024llama33modelcard,dubey2024llama}, BF16, temperature 0.0) is evaluated 
under the same k-penalty protocol over $k=1..5$ on the 150-dialogue held-out 
split from the \texttt{mturk\_agent\_1} perspective ($n=746$ k-prefix predictions, 
parse rate 95.7\%); single-split evaluation, so per-fold CV variance is not reported. The supervised SFT baseline is a 
Llama-3.1-8B ~\citep{meta2024llama31modelcard,dubey2024llama} model fine-tuned via LoRA; it is a same-task supervised 
reference for opponent-priority prediction at 8B scale, distinct from 
the Bayesian teacher and distilled student evaluated in 
Sections~\ref{sec:results:calibration} and \ref{sec:results:decision}.
}
\label{tab:opponent_comparison}
\end{table}

The remaining experiments evaluate whether our Bayesian framework --- which 
additionally exposes a calibrated turn-level posterior 
(Section~\ref{sec:results:calibration}) --- preserves opponent-modeling 
capability while supporting auditable belief tracking that the prompted 
70B baseline cannot provide.

\FloatBarrier
\section{Deal-or-No-Deal Transfer Details}
\label{app:dnd_transfer}

\begin{table}[H]
\centering
\scriptsize
\setlength{\tabcolsep}{3pt}
\begin{tabular}{lccccccc}
\toprule
Run & DND Brier $\downarrow$ & EMA@2 $\uparrow$ & EMA$_{1:3}$ $\uparrow$ & NDCG$_{1:3}$ $\uparrow$ & Bid cos. $\uparrow$ & Self-util. $\uparrow$ & $n$ \\
\midrule
Direct zero-shot native & 0.3303 & 0.1728 & 0.1672 & 0.5013 & 0.7092 & 0.7619 & 1695 \\
Direct zero-shot renamed & 0.3190 & 0.1806 & 0.1743 & 0.5170 & 0.7191 & 0.7910 & 1695 \\
Few-shot native empirical & \textbf{0.1166} & \textbf{0.5981} & \textbf{0.5879} & \textbf{0.8541} & \textbf{0.8380} & 0.7263 & 1695 \\
Few-shot renamed 5/4/3 & 0.1184 & 0.5786 & 0.5713 & 0.8486 & 0.7391 & 0.7598 & 1695 \\
Few-shot renamed empirical & 0.1207 & 0.5845 & 0.5694 & 0.8459 & 0.8332 & 0.7259 & 1695 \\
Rule baseline & 0.1605 & 0.2369 & 0.2259 & 0.5675 & 0.7184 & 0.7842 & 1695 \\
Zero-shot native & 0.1438 & 0.4796 & 0.4827 & 0.7852 & 0.7374 & 0.7576 & 1695 \\
Zero-shot renamed & 0.1335 & 0.5126 & 0.5029 & 0.7969 & 0.7345 & 0.7612 & 1695 \\
\bottomrule
\end{tabular}
\caption{
Full DND transfer results. Brier, EMA, and NDCG are computed on strict partner orderings. 
Bid cosine compares the top planned allocation with the final human allocation when available. 
Self-util. is the normalized self utility of the top planned allocation, not final negotiation success. DND Brier uses the sum-normalized multiclass definition, with six-way uniform reference $1/6 \approx 0.167$. This normalization differs from the CaSiNo class-mean Brier used in the main paper, whose six-way uniform reference is $5/36 \approx 0.139$; DND and CaSiNo Brier values should therefore be compared only within their respective sections.
}
\label{tab:dnd_full}
\end{table}

\FloatBarrier
\section{DND transfer protocol.}
For cross-domain transfer, we evaluate on Deal-or-No-Deal using the same six-ordering belief interface, replacing CaSiNo issues with DND item types. Since DND value vectors may contain ties, headline belief metrics are computed only on strict partner orderings. DND Brier uses the sum-normalized multiclass definition, $\sum_k(p_k-y_k)^2/(K-1)$, for which the six-way uniform reference is $1/6$. We report EMA@2 and k-penalized EMA/NDCG over $k=1$--3 because DND dialogues are shorter and later-turn support drops quickly.

\newpage

\end{document}